\documentclass{article} 
\usepackage{iclr2027_conference,times}

\usepackage{amsmath,amsfonts,bm}

\def\eqref#1{equation~\ref{#1}}

\def\1{\bm{1}}

\DeclareMathAlphabet{\mathsfit}{\encodingdefault}{\sfdefault}{m}{sl}
\SetMathAlphabet{\mathsfit}{bold}{\encodingdefault}{\sfdefault}{bx}{n}

\usepackage{xcolor}
\definecolor{deeppink}{rgb}{1.0, 0.08, 0.58}

\usepackage{url}
\usepackage{graphicx}
\usepackage{multirow}
\usepackage{booktabs}
\usepackage{threeparttable}
\usepackage{tabularx}
\usepackage{float}
\usepackage{algorithm,algorithmic}
\usepackage{amsmath,amssymb,amsthm}
\usepackage{capt-of}

\usepackage{wrapfig}
\usepackage{placeins}

\usepackage{hyperref}
\hypersetup{
    colorlinks=true,
    linkcolor=red,
    urlcolor=deeppink,
    citecolor=blue,
    pdfauthor={Zhaoyu Chen and Yongwei Nie}
}

\newcommand{\method}{\textsc{DTP-OPSD}}

\title{DTP-OPSD: Reinforcement Learning with Distractor-Aware Temporal-Privilege On-Policy Self-Distillation for Temporal Video Grounding}

\author{
Zhaoyu Chen \quad Yongwei Nie \\
School of Computer Science and Engineering \\
South China University of Technology, Guangzhou, China
}

\iclrfinalcopy 

\begin{document}

\maketitle
\lhead{Preprint}

\begin{abstract}
Temporal Video Grounding (TVG) aims to localize the start and end times of an event described by a language query in a video. The task becomes particularly challenging when the same video contains multiple repeated, similar, or opposite events. On these hard examples, all responses sampled by Group Relative Policy Optimization (GRPO) may miss the target and receive zero temporal Intersection-over-Union (tIoU). Their group-relative advantages are then all zero, leaving no useful policy gradient signal. Further analysis shows that the model is often able to recognize the queried event but is distracted by other events in the same video and selects the wrong occurrence. To address this, we propose DTP-OPSD, a reinforcement learning framework with distractor-aware temporal-privilege on-policy self-distillation. The student observes the clean video and generates responses on-policy, while a shared-weight teacher observes a privileged video in which the target window is marked by red full-frame borders and same-video distractor windows by blue full-frame borders. For the distillation loss, the teacher evaluates student-generated prefixes and provides dense token-level Jensen--Shannon divergence supervision. This supervision remains informative even when GRPO provides no policy gradient signal. We suppress distillation when the teacher does not outperform the student and strengthen it for groups where the teacher performs better but all student rollouts receive zero tIoU. We also construct a compact hard-interference training set through VLM screening and human verification. With only 300 additional post-training examples, DTP-OPSD achieves
the highest mIoU among the compared open-source 8B models on
Charades-TimeLens, ActivityNet-TimeLens, and QVHighlights-TimeLens,
reaching 55.93, 54.77, and 66.71, respectively. Compared with Pure
GRPO trained on the same data, DTP-OPSD improves all 12 reported
metrics, with R1@0.7 gains of 1.28, 1.62, and 1.78 points on the three
benchmarks. Code will be released at
\url{https://github.com/nian1125/DTP-OPSD}.
\end{abstract}

\section{Introduction}
Temporal Video Grounding (TVG) aims to identify the start and end
times of an event described by a language query
\citep{Gao2017TALL,Hendricks2017MCN}.
It is a fundamental video-language task that requires joint
understanding of the query semantics and the temporal structure of
the video. Earlier methods localize target moments through proposal
matching, boundary regression, or span prediction
\citep{Zhang2020VSLNet,Lei2021QVHighlights}.
More recently, multimodal large language models have formulated TVG
as timestamp generation through a language interface
\citep{Ren2024TimeChat,Huang2024VTimeLLM,%
Wang2024GroundedVideoLLM,Zhang2026TimeLens}.
This formulation makes TVG more flexible, but accurate localization
remains difficult when a video contains multiple related event
occurrences that provide plausible matches to the query.

Reinforcement learning is now widely used to improve TVG with
multimodal large language models
\citep{Wang2025TimeR1,Chen2025TVGRL,Dong2025VideoTGR1}.
A predicted temporal window can be scored directly against the
annotated target using temporal Intersection-over-Union (tIoU). The tIoU score provides a direct reward, without requiring a learned
reward model or extra reward labels. Group Relative Policy Optimization
(GRPO) samples several responses for the same input and trains the policy
using their relative rewards within the group
\citep{Shao2024DeepSeekMath,DeepSeekAI2025R1}.
However, GRPO cannot learn from a rollout group when all sampled
responses receive the same reward, because mean centering assigns
every response zero relative advantage. In TVG, a particularly
consequential case is the all-zero group.

A major challenge arises when a video contains multiple events related
to the query. The queried event may occur repeatedly, while other
moments may contain visually or semantically similar actions or
opposite actions involving the same actor and object. These distractors
make it difficult to identify the specific occurrence requested by the
query. We refer to this problem as same-video interference. Under
strong interference, the model may recognize the queried event but
still select the wrong occurrence.

\begin{figure*}[t]
\centering
\includegraphics[width=\textwidth]{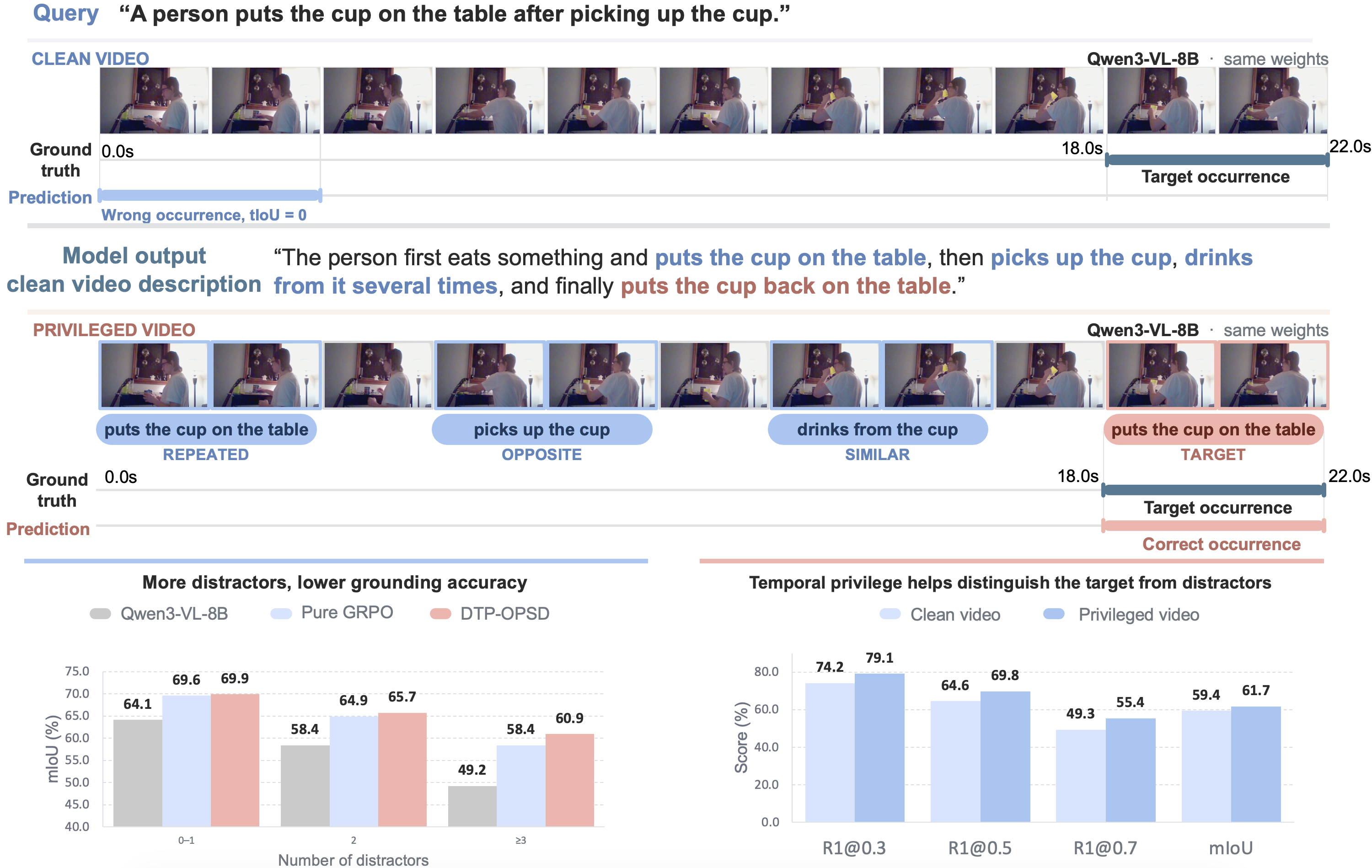}
\caption{
\textbf{Top.}
Qwen3-VL-8B recognizes the relevant actions but selects an earlier
occurrence from the clean video, resulting in zero tIoU. The
privileged teacher view marks the target with red borders and the
distractors with blue borders, helping the same model recover the
correct interval.
\textbf{Bottom left.}
On QVHighlights-TimeLens, mIoU decreases as the number of valid
distractors increases, while DTP-OPSD degrades less than
Qwen3-VL-8B and Pure GRPO.
\textbf{Bottom right.}
On the full QVHighlights-TimeLens evaluation set, the privileged view improves localization over the clean view under identical model weights.
}
\label{fig:motivation}
\end{figure*}

Figure~\ref{fig:motivation} illustrates this failure. In this example, Qwen3-VL-8B~\citep{Bai2025Qwen3VL} recognizes the
queried action but selects an earlier occurrence, resulting in zero
tIoU. With the same model weights, the privileged view helps distinguish
the target from distractors and recover the correct interval.
Dataset-level diagnostics further show that localization becomes more
difficult as the number of valid same-video distractors increases,
while DTP-OPSD remains more robust. The privileged view also outperforms the clean view on the full
QVHighlights-TimeLens evaluation set. These results suggest that some failures arise from
incorrect occurrence selection under same-video interference rather
than from a failure to recognize the queried action.

Under strong same-video interference, all sampled responses may select
distractor occurrences and receive zero tIoU, leaving some of the
hardest examples inactive in the current GRPO update.

To reduce wrong-occurrence predictions and provide a learning signal for
all-zero rollout groups,
we propose DTP-OPSD, a reinforcement learning framework with
distractor-aware temporal-privilege on-policy self-distillation.
As shown in Figure~\ref{fig:method_overview}, the student observes the
clean video and samples responses on-policy. A teacher sharing the
student's current weights observes a training-only privileged view,
where the target region is marked in red and the distractors in blue.
For distillation, the teacher evaluates student-generated prefixes and
provides dense token-level Jensen--Shannon divergence supervision
\citep{Agarwal2024GKD,Zhao2026OPSD,Yuan2026VisionOPD}.

Since the privileged teacher is not reliable on every sample, we
introduce an adaptive distillation controller. The controller compares the mean Student
tIoU from the GRPO rollouts with the mean teacher tIoU from
independently generated teacher rollouts under the same sampling
settings. It suppresses distillation when the teacher does not
outperform the student and strengthens it for groups where the teacher
performs better but all student rollout tIoUs are zero. All remaining
groups use the default distillation weight. This design prevents
unreliable teacher supervision while providing stronger guidance when
GRPO receives no reward-driven learning signal.

Effective self-distillation requires hard but teachable examples. We further construct \textsc{HI-TVG-300} from the TimeLens-100K
training set~\citep{Zhang2026TimeLens}, without introducing external
videos. VLM screening and human verification identify examples with
strong same-video interference. A teachability filter then retains
examples where the privileged teacher clearly outperforms the
clean-video student. The resulting dataset contains 300 high-quality,
difficult but teachable training examples.
Our contributions are:
\begin{itemize}
\item We study a key challenge in reinforcement learning-based
TVG. Same-video distractors can cause wrong-occurrence predictions
and all-zero tIoU groups that provide no relative learning signal
to GRPO.
\item We propose distractor-aware temporal privilege, using red
borders to mark the target region and blue borders to mark
distractor regions. This enables on-policy distillation from a
shared-weight teacher to a clean-video student.
\item We design an adaptive distillation controller that suppresses
distillation when the teacher does not outperform the student and
strengthens supervision for all-zero student rollout groups when the
teacher performs better.
\item We construct \textsc{HI-TVG-300}, a high-quality hard-interference
training set, through VLM screening, human verification, and
teacher--student teachability filtering.
\end{itemize}

\section{Related Work}
\paragraph{Temporal video grounding and reinforcement learning.}
Early TVG methods localize moments through proposal matching, boundary
regression, or span prediction
\citep{Gao2017TALL,Hendricks2017MCN,Zhang2020VSLNet,
Lei2021QVHighlights}.
Video MLLMs later introduced timestamp-aware representations and
grounding instruction tuning for generative temporal localization
\citep{Ren2024TimeChat,Huang2024VTimeLLM,Huang2024LITA,
Wang2024GroundedVideoLLM,Zeng2025TimeSuite,
Guo2026T2SGrid,Guo2025TAR}.
Recent work extends TVG to multi-moment annotations
\citep{Cao2025QVM2,Xu2026OMTG}.
Reinforcement learning with verifiable rewards has also been applied
to multimodal and video reasoning
\citep{Huang2025VisionR1,Feng2025VideoR1,Li2025VideoChatR1},
and more recently to temporal video grounding
\citep{Wang2025TimeR1,Chen2025TVGRL,Chen2026Invert4TVG,
Dong2025VideoTGR1,Zhang2026TimeLens}.
However, group-relative optimization loses its relative learning
signal when all rollout rewards are identical. We study this failure
under same-video interference, where competing occurrences can produce
wrong-occurrence predictions and all-zero tIoU groups.

\paragraph{On-policy distillation and self-distillation.}
Knowledge distillation transfers a teacher distribution to a student
\citep{Hinton2015KD}. On-policy distillation provides teacher
supervision along student-generated trajectories, while On-Policy
Self-Distillation creates a teacher--student capability gap by
conditioning shared model weights on different inputs
\citep{Agarwal2024GKD,Zhao2026OPSD}.
Recent multimodal self-distillation methods construct privileged views
from localized visual evidence or higher-resolution images
\citep{Yuan2026VisionOPD,Zhu2026RPOPSD}.
Video-OPD applies on-policy distillation to temporal video grounding
using a fixed frontier teacher to supervise student-generated
trajectories \citep{Li2026VideoOPD}.
In contrast, DTP-OPSD combines GRPO with shared-weight
self-distillation and constructs query-dependent temporal privilege by
marking the target occurrence and its same-video distractors. Its
adaptive controller suppresses unreliable teacher supervision and
strengthens it for all-zero student groups.

\section{Method}
\label{sec:method}
\begin{figure*}[t]
\centering
\includegraphics[width=\textwidth]{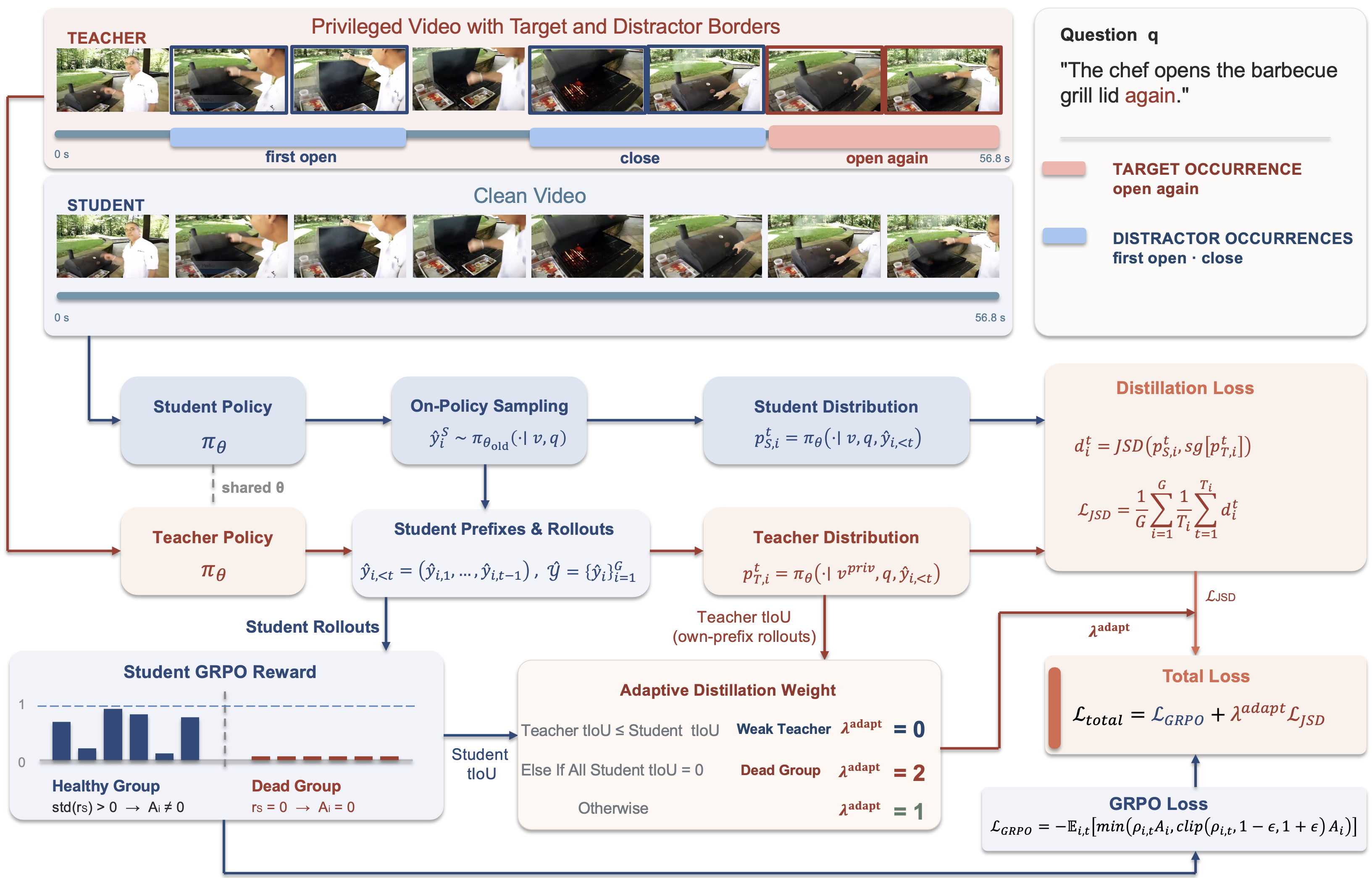}
\caption{
Overview of \method{}. The clean-video student samples $G$ on-policy
responses, whose tIoUs provide the GRPO rewards. The shared-weight
teacher observes the privileged video and evaluates student-generated
prefixes for token-level JSD. Separately, mean teacher tIoU is
estimated from independent own-prefix rollouts under the same sampling
settings. The controller uses the teacher and student tIoUs to adjust
the group-level distillation weight. For clarity, the figure shows the
group-level adaptive weighting, while
Equation~\ref{eq:full_objective} gives the complete minibatch
objective.
}
\label{fig:method_overview}
\end{figure*}

Given a video $V$ and a language query $q$, the model predicts a
temporal interval $\hat{\tau}=[\hat{t}_s,\hat{t}_e]$, with the
ground-truth interval denoted by $\tau=[t_s,t_e]$. As shown in
Figure~\ref{fig:method_overview}, \method{} combines GRPO with
distractor-aware temporal-privilege on-policy self-distillation.
Section~\ref{sec:grpo_tvg} describes tIoU-based GRPO and the all-zero
rollout groups caused by same-video interference.
Section~\ref{sec:dtp_opsd} introduces temporal privilege and
same-prefix JSD supervision. Section~\ref{sec:adaptive_control}
presents teacher competence estimation and adaptive distillation
control. Finally, Section~\ref{sec:data_construction} describes the
construction of \textsc{HI-TVG-300}. In the notation below, $g$ indexes a video--query rollout group
$(V_g,q_g)$, while $i$ indexes one of the $G$ student responses sampled
for that group.

\subsection{GRPO for TVG and Dead Groups}
\label{sec:grpo_tvg}

\paragraph{Reward and group-relative optimization.}
Following the thinking-free RLVR setting of
TimeLens~\citep{Zhang2026TimeLens}, our only reward is temporal
Intersection-over-Union:

\begin{equation}
\operatorname{tIoU}(\hat{\tau},\tau)
=
\frac{|\hat{\tau}\cap\tau|}
{|\hat{\tau}\cup\tau|}.
\label{eq:tiou_reward}
\end{equation}

Here, $|\cdot|$ denotes the duration of an interval. A response that
cannot be parsed into a valid interval receives zero reward. Unlike
Time-R1~\citep{Wang2025TimeR1}, we use no format or temporal-distance
reward, keeping the training reward consistent with the standard TVG
evaluation metric.

For each video-query pair $(V_g,q_g)$, the rollout policy
$\pi_{\theta_{\mathrm{old}}}$ samples $G$ responses conditioned on the
clean video and query. Each response is parsed into a predicted
interval $\hat{\tau}_{g,i}$, and its tIoU is used as the student reward
$r_{g,i}^{S}$. Following TimeLens, we compute the group-relative
advantages by mean centering without standard-deviation scaling:

\begin{equation}
\bar{r}_g^{S}
=
\frac{1}{G}\sum_{j=1}^{G}r_{g,j}^{S},
\qquad
A_{g,i}
=
r_{g,i}^{S}-\bar{r}_g^{S}.
\label{eq:group_advantage}
\end{equation}

We use these advantages in the group-relative GRPO objective
\citep{Shao2024DeepSeekMath,Zhang2026TimeLens}, denoted by
$\mathcal{L}_{\mathrm{GRPO}}$. The exact loss,
completion-token normalization, and single-update optimization
procedure are provided in Appendix~\ref{app:training_algorithm}.

\paragraph{Dead groups under same-video interference.}
GRPO obtains a reward-driven learning signal from a rollout group only
when its rewards differ. Under strong same-video interference, a
critical failure mode occurs when every sampled response selects an
incorrect occurrence that does not overlap with the target. In this
case, $r_{g,i}^{S}=0$ for all $i$, and
Equation~\ref{eq:group_advantage} yields $A_{g,i}=0$ for all $i$.
The group therefore contributes no reward-driven policy gradient to
the current update. We refer to such an all-zero rollout group as a
dead group. Although any constant-reward group has zero relative
advantages, we use the all-zero condition to identify dead groups
because it indicates complete failure to overlap the target. When the
privileged teacher achieves a higher mean tIoU than the student,
\method{} provides strengthened token-level supervision to these
groups without relying on within-group reward differences.

\subsection{Distractor-Aware Temporal-Privilege On-Policy Self-Distillation}
\label{sec:dtp_opsd}

\paragraph{Distractor-aware temporal privilege.}
For each training example, we construct an aligned clean video $V$
and a privileged video $V^{\mathrm{priv}}$ with identical content and
duration. In $V^{\mathrm{priv}}$, frames inside the target window
receive red full-frame borders, while frames inside the verified
same-video distractor windows receive blue full-frame borders. Before
rendering, we expand only the target window to $1.5$ times its duration
around its center. The red borders provide a coarse GT-derived temporal cue rather than
displaying the annotated boundaries verbatim. Red takes priority wherever
the target and distractor windows overlap. The clean video remains
unchanged, and $V^{\mathrm{priv}}$ is used only during training.

\paragraph{Shared-weight teacher and student.}
Following On-Policy Self-Distillation
(OPSD)~\citep{Zhao2026OPSD}, the teacher and student are two
conditional views of the same model. They share the current parameters
$\theta$ and use the same text prompt, which does not mention the
border colors. Their conditioning differs only in the video input.
The resulting capability gap therefore comes from the privileged
temporal evidence rather than additional parameters or textual hints.

\paragraph{On-policy same-prefix distillation.}
The student responses sampled for GRPO are reused for distillation. At
each token position, the student and teacher compute their next-token
distributions from the clean and privileged videos, respectively,
while conditioning on the same student-generated prefix
$\hat y_{g,i,<t}$. We denote these distributions by
$p_{S,g,i}^{t}$ and $p_{T,g,i}^{t}$. The teacher therefore evaluates
the generation states actually visited by the student, keeping the
supervision on-policy.

We align the two distributions using Jensen--Shannon divergence
(JSD)~\citep{Lin1991JSD}:

\begin{equation}
\begin{aligned}
d_{g,i}^{t}
&=
\operatorname{JSD}\!\left(
p_{S,g,i}^{t},
\operatorname{sg}\!\left[p_{T,g,i}^{t}\right]
\right),\\
\ell_g^{\mathrm{JSD}}
&=
\frac{1}{G}\sum_{i=1}^{G}
\frac{1}{T_{g,i}}
\sum_{t=1}^{T_{g,i}}d_{g,i}^{t}.
\end{aligned}
\label{eq:jsd}
\end{equation}

Here, $\operatorname{sg}[\cdot]$ denotes stop-gradient, and $T_{g,i}$
is the number of valid completion tokens. Gradients propagate only
through the clean-video student branch. This loss does not vanish
merely because all student rollout tIoUs are zero.

\subsection{Adaptive Distillation Control}
\label{sec:adaptive_control}

\paragraph{Teacher competence from independent rollouts.}
The privileged view does not guarantee a better teacher prediction for
every input. The teacher may ignore the border information or still
select a distractor, so unreliable teacher supervision should be
suppressed. To estimate teacher competence fairly, the teacher uses
the same model snapshot, text prompt, sampling temperature, and number
of rollouts $G$ as the student. The only difference is the video input.
The teacher independently generates $G$ responses from the privileged
video using its own autoregressive prefixes. Each response is parsed
using the same rule as the student responses, and an invalid response
receives zero reward. We denote the mean tIoU of these responses by
$\bar r_g^{T}$ and use it as the teacher competence estimate. These
teacher rollouts are used only by the adaptive controller and do not
enter the JSD loss, which remains defined on student-generated
prefixes in Equation~\ref{eq:jsd}.

\paragraph{Adaptive distillation weight.}
Let $\bar r_g^{S}$ denote the mean student tIoU defined in
Equation~\ref{eq:group_advantage}. The group-level distillation
multiplier is

\begin{equation}
\lambda_g^{\mathrm{adapt}}
=
\begin{cases}
0,
& \bar r_g^{T}\leq\bar r_g^{S},\\[2pt]
2,
& \bar r_g^{T}>\bar r_g^{S}
\ \text{and}\ 
r_{g,i}^{S}=0\ \ \forall i,\\[2pt]
1,
& \text{otherwise}.
\end{cases}
\label{eq:quality_controller}
\end{equation}

The first condition removes distillation when the privileged teacher
does not achieve a higher mean tIoU than the student. This includes
groups where both the student and teacher receive only zero tIoU. The
second condition strengthens distillation only when the teacher
performs better and all student rollout tIoUs are zero. When the
teacher performs better, all remaining groups, including student
groups with identical nonzero rewards, retain the default weight.

\paragraph{Final objective.}
The overall objective combines the standard GRPO loss with
group-adaptive token-level distillation:

\begin{equation}
\mathcal{L}_{\mathrm{total}}
=
\mathcal{L}_{\mathrm{GRPO}}
+
\lambda_{\mathrm{JSD}}
\frac{1}{|\mathcal{B}|}
\sum_{g\in\mathcal{B}}
\lambda_g^{\mathrm{adapt}}
\ell_g^{\mathrm{JSD}},
\label{eq:full_objective}
\end{equation}

where $\mathcal{B}$ is a minibatch of rollout groups and
$\lambda_{\mathrm{JSD}}$ is the global distillation coefficient. The
GRPO coefficient and $\lambda_{\mathrm{JSD}}$ are both set to $1$ in
our final configuration. The group-level multiplier
$\lambda_g^{\mathrm{adapt}}\in\{0,1,2\}$ controls the contribution of
JSD for each rollout group. It removes unreliable teacher supervision
and strengthens supervision for all-zero student groups when the
teacher demonstrates better localization.
\subsection{Construction of HI-TVG-300}
\label{sec:data_construction}

\begin{figure*}[t]
    \centering
    \includegraphics[
    width=0.97\textwidth,
    keepaspectratio
]{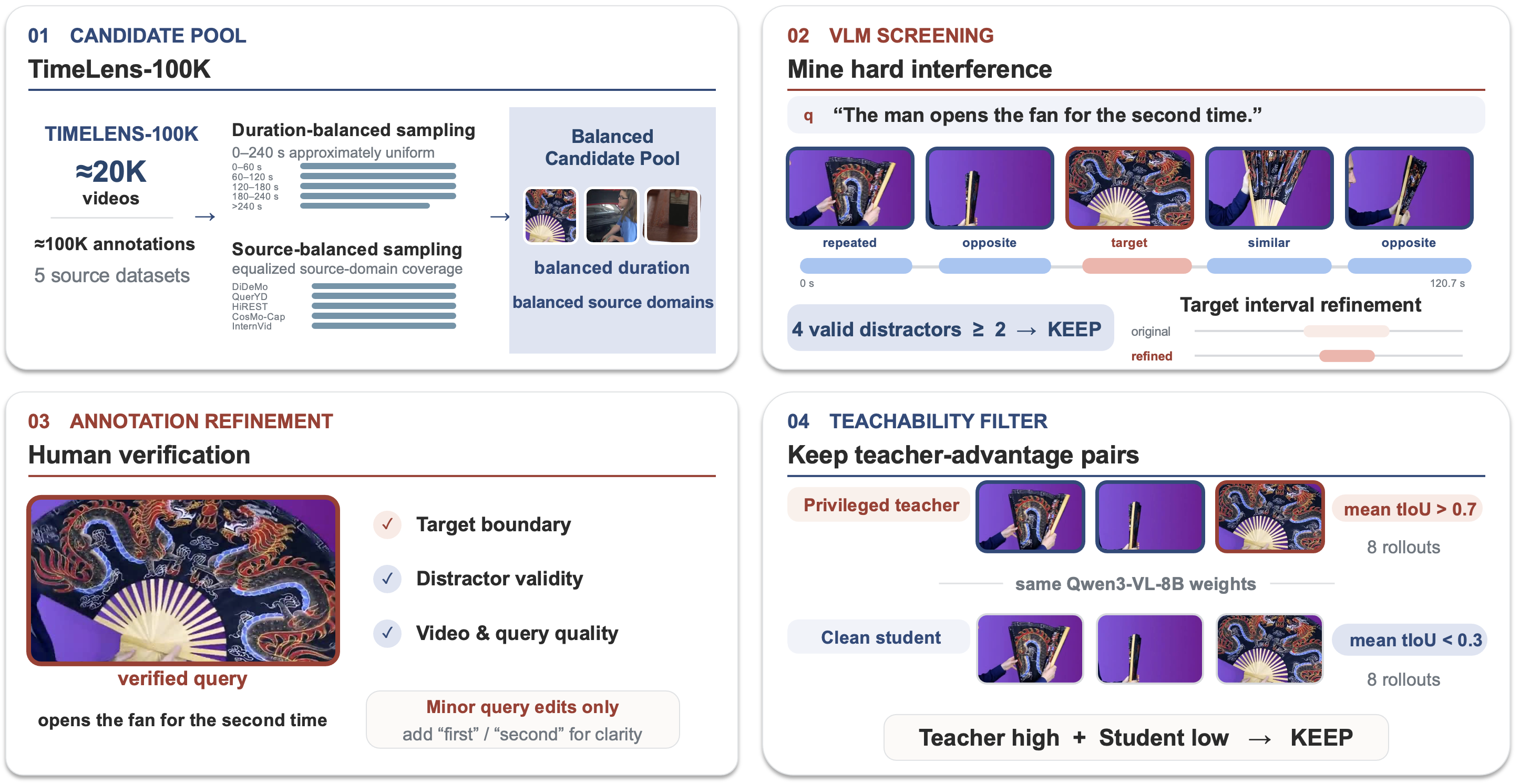}
    \caption{
    Construction pipeline of \textsc{HI-TVG-300}, derived entirely
    from the TimeLens-100K training set. Candidate videos are balanced
    by duration and source, mined for same-video interference, verified
    by human annotators, and filtered according to the privileged
    teacher advantage.
    }
    \label{fig:data_construction}
\end{figure*}

Large-scale TVG datasets provide broad training coverage but are not
designed to contain strong same-video interference or reliable
privileged supervision. We therefore derive \textsc{HI-TVG-300}
entirely from the TimeLens-100K training set
\citep{Zhang2026TimeLens}. It reuses selected videos and queries while
refining target intervals and adding verified distractor intervals,
without introducing external videos or queries. As summarized in Figure~\ref{fig:data_construction}, we first balance
candidate videos across duration ranges and source datasets. GPT-5.6
then examines each complete video and its annotations to identify
repeated, similar, or opposite event occurrences and propose refined
target intervals. Human annotators verify the target boundaries,
distractor intervals, video quality, and queries. We retain examples
containing at least two valid same-video distractors. Finally, using
the same Qwen3-VL-8B checkpoint and sampling settings, the clean-video
student and privileged teacher independently generate eight responses.
We retain an example when the mean teacher tIoU exceeds $0.7$ and the
mean student tIoU is below $0.3$. This process produces 300 difficult
but teachable examples with verified same-video interference. Further construction details, annotation protocols, and dataset
statistics are provided in
Appendix~\ref{sec:data_stats}.

\section{Experiments}
\label{sec:experiments}

We evaluate \method{} through comparisons with representative TVG
methods and controlled ablations, with additional qualitative examples
provided in Appendix~\ref{sec:qualitative_examples}.

\subsection{Experimental Setup}
\label{sec:experimental_setup}

\paragraph{Benchmarks and evaluation metrics.}
We evaluate \method{} on Charades-STA~\citep{Gao2017TALL},
ActivityNet Captions~\citep{Krishna2017ActivityNet}, and
QVHighlights~\citep{Lei2021QVHighlights}. Following
TimeLens~\citep{Zhang2026TimeLens}, we use its corrected benchmarks
for analysis and final reporting,
referred to as Charades-TimeLens, ActivityNet-TimeLens, and
QVHighlights-TimeLens. We further evaluate on
OMTG-Bench~\citep{Xu2026OMTG} and
QV-M$^{2}$~\citep{Cao2025QVM2}, where a query may correspond to
multiple valid temporal intervals. We report mIoU as the primary
metric, together with R1@0.3, R1@0.5, and R1@0.7. mIoU averages the
per-example tIoU, while R1@$m$ measures the percentage of predictions
with a tIoU of at least $m$.

\paragraph{Implementation details.}
We initialize Qwen3-VL-8B from the publicly released TimeLens
supervised checkpoint, denoted by TimeLens-SFT, and
post-train it on \textsc{HI-TVG-300}, which is derived entirely from
the TimeLens-100K training set. Videos are sampled at 2 FPS with a
maximum of 448 frames and a visual-token budget of 14,336. For each
input, the student samples $G=8$ responses for GRPO. The teacher
independently samples the same number of responses under identical
sampling settings to estimate teacher competence. These teacher
rollouts are not used to compute the JSD loss. We use a global batch of eight prompt-level rollout groups and a
learning rate of $1\times10^{-6}$. Unless otherwise specified, results from our post-training experiments
and training ablations are reported as the mean over three independent
training runs. The global JSD coefficient is set to
$\lambda_{\mathrm{JSD}}=1$, and the reference-policy KL coefficient is
set to $\beta=0$. Additional training and implementation details are provided in
Appendix~\ref{app:implementation_details}.

\subsection{Comparison with State-of-the-Art Methods}
\label{sec:sota_comparison}

\paragraph{Main results.}
Table~\ref{tab:sota_tvg} compares \method{} with representative
open-source TVG models on three TimeLens benchmarks. Pure GRPO trained
on \textsc{HI-TVG-300} already improves the TimeLens-SFT initialization
on all three benchmarks. Building on this strong baseline, \method{}
achieves 55.93, 54.77, and 66.71 mIoU on Charades-TimeLens,
ActivityNet-TimeLens, and QVHighlights-TimeLens, respectively. These
are the highest mIoU scores among the compared open-source 8B models.
Compared with Pure GRPO, \method{} improves all 12 reported metrics,
including R1@0.7 gains of 1.28, 1.62, and 1.78 points on the three
benchmarks. Among all compared models, \method{} ranks first on 11 of
the 12 metrics and second on the remaining metric, demonstrating
consistent improvements across different overlap thresholds.

\begin{table*}[t]
\centering
\caption{
Comparison with representative open-source TVG models on three
TimeLens benchmarks. We evaluate all model checkpoints ourselves using the corrected
TimeLens annotations and the same TimeLens evaluation protocol.
TimeLens-SFT provides the common initialization for TimeLens-8B,
Pure GRPO, and DTP-OPSD. Both Pure GRPO and DTP-OPSD are post-trained
on our \textsc{HI-TVG-300} set. The best and second-best results are
shown in bold and underlined, respectively.
}
\label{tab:sota_tvg}

\scriptsize
\setlength{\tabcolsep}{2.0pt}
\renewcommand{\arraystretch}{1.0}

\begin{tabular*}{\textwidth}{
@{\extracolsep{\fill}}l*{12}{c}@{}
}
\toprule
& \multicolumn{4}{c}{\textbf{Charades-TimeLens}}
& \multicolumn{4}{c}{\textbf{ActivityNet-TimeLens}}
& \multicolumn{4}{c}{\textbf{QVHighlights-TimeLens}} \\
\cmidrule(lr){2-5}
\cmidrule(lr){6-9}
\cmidrule(lr){10-13}

\textbf{Method}
& R1@0.3 & R1@0.5 & R1@0.7 & mIoU
& R1@0.3 & R1@0.5 & R1@0.7 & mIoU
& R1@0.3 & R1@0.5 & R1@0.7 & mIoU \\
\midrule

VideoChat-R1-7B
& 51.93 & 30.86 & 11.72 & 33.78
& 35.03 & 23.94 & 11.36 & 25.01
& 29.37 & 19.13 & 9.42 & 21.56 \\

Time-R1-7B
& 57.95 & 32.07 & 16.96 & 36.62
& 44.84 & 31.08 & 19.02 & 33.19
& 65.87 & 51.59 & 36.14 & 49.26 \\

MiMo-VL-7B
& 57.91 & 42.63 & 20.58 & 39.65
& 49.34 & 38.76 & 22.48 & 35.52
& 57.12 & 42.61 & 28.46 & 41.57 \\

Qwen3-VL-8B
& 69.24 & 53.41 & 27.57 & 48.32
& 62.18 & 51.29 & 34.43 & 46.86
& 74.27 & 64.69 & 49.35 & 59.48 \\

Video-OPD-8B
& 75.62 & 60.57 & 32.71 & 53.16
& 66.19 & 57.13 & 39.47 & 51.82
& 76.94 & 68.72 & 54.45 & 64.39 \\

TimeLens-8B
& 76.61 & 63.03 & \underline{35.26} & 55.28
& \underline{68.97} & 58.45 & 40.68 & 53.23
& \underline{80.21} & \textbf{71.62} & 55.57 & 65.59 \\

\midrule

TimeLens-SFT
& 74.19 & 59.35 & 31.12 & 51.85
& 63.44 & 54.09 & 36.61 & 49.26
& 69.57 & 59.31 & 45.38 & 57.34 \\

Pure GRPO
& \underline{77.38} & \underline{64.17} & 35.15
& \underline{55.43}
& 68.26 & \underline{59.58} & \underline{41.85}
& \underline{53.62}
& 79.75 & 70.93 & \underline{55.61}
& \underline{65.95} \\

\method{}
& \textbf{77.94} & \textbf{64.67} & \textbf{36.43}
& \textbf{55.93}
& \textbf{69.67} & \textbf{60.71} & \textbf{43.47}
& \textbf{54.77}
& \textbf{80.53} & \underline{71.45} & \textbf{57.39}
& \textbf{66.71} \\

\bottomrule
\end{tabular*}
\end{table*}

\begin{figure*}[tbp]
\centering
\includegraphics[
    width=0.97\textwidth,
    keepaspectratio
]{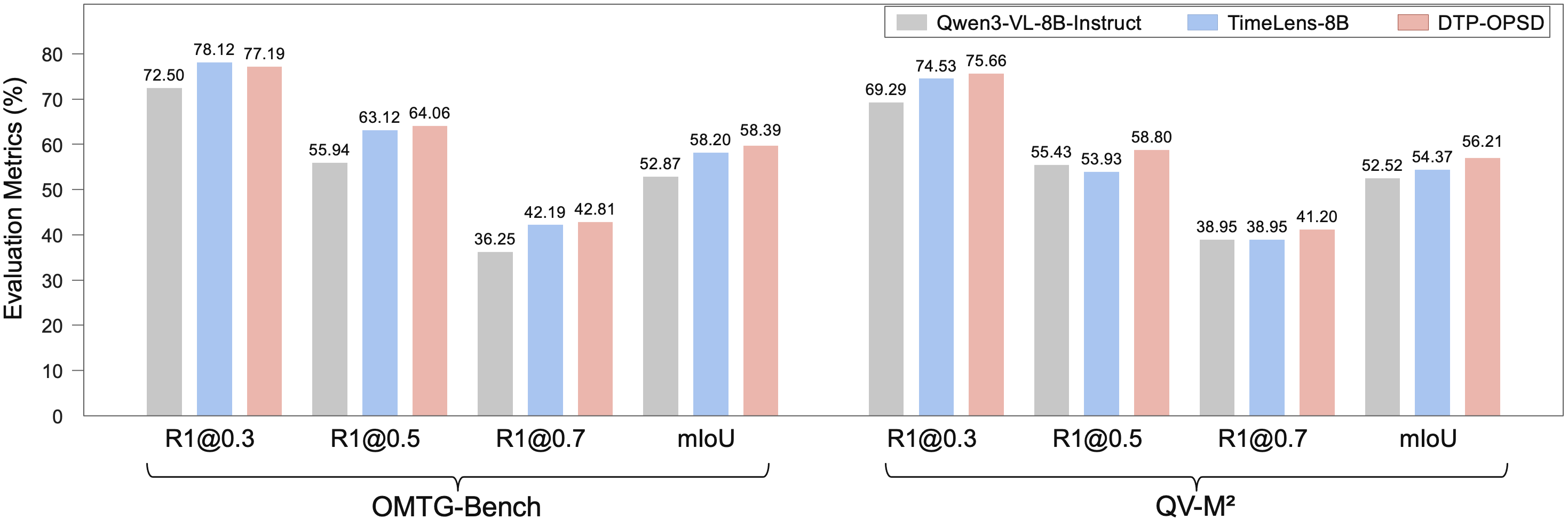}
\caption{
Single-interval grounding on OMTG-Bench and QV-M$^{2}$, both with
multi-moment annotations. Each model predicts one interval
and is evaluated against its best-matching ground-truth interval.
Results are reported on all 320 OMTG-Bench queries and all 323
QV-M$^{2}$ test queries.
}
\label{fig:multi_moment_results}
\end{figure*}

\paragraph{Single-interval grounding under multi-moment annotations.}
We further evaluate \method{} on OMTG-Bench~\citep{Xu2026OMTG} and
QV-M$^{2}$~\citep{Cao2025QVM2}. All models retain the standard
single-interval prompt and output format, without benchmark-specific
training or multi-interval decoding. For both benchmarks, we adopt the
mIoU@1 protocol used by QV-M$^{2}$. Each prediction is evaluated
against its best-matching ground-truth interval, and the resulting
tIoU scores are used to compute mIoU and R1. As shown in
Figure~\ref{fig:multi_moment_results}, \method{} achieves 58.39 mIoU
on OMTG-Bench and 56.21 mIoU on QV-M$^{2}$. It performs comparably
with TimeLens-8B on OMTG-Bench and exceeds it by 1.84 points on
QV-M$^{2}$. These results show that \method{} remains effective when
a query has multiple valid temporal occurrences. This protocol
evaluates whether the model identifies one valid occurrence rather
than whether it retrieves all annotated intervals.

\subsection{Ablation Study}
\label{sec:ablation}

We ablate the adaptive controller, privileged video design, and data
curation, with additional results in
Appendix~\ref{app:additional_ablations}. All variants use Qwen3-VL-8B
initialized from TimeLens-SFT, with the same training settings and
evaluation step. Except for the data ablation, all variants
use \textsc{HI-TVG-300}. These analyses determine one shared configuration, which is applied
unchanged to all other benchmarks.

\paragraph{Effect of the adaptive distillation controller.}
Table~\ref{tab:controller_ablation} evaluates adaptive JSD weighting.
All variants retain GRPO and privileged-view JSD. Fixed JSD sets
$\lambda_g^{\mathrm{adapt}}=1$ for every group, whereas Pure GRPO in
Table~\ref{tab:sota_tvg} removes JSD entirely. Rescue only sets
$\lambda_g^{\mathrm{adapt}}=2$ for eligible all-zero groups and
otherwise retains the default weight, while Suppression only sets
$\lambda_g^{\mathrm{adapt}}=0$ when the teacher does not outperform
the student. Both variants improve over Fixed JSD, and their combination
achieves the best mIoU on all three benchmarks, with gains of 0.45,
0.75, and 0.93 points. The full controller consistently outperforms both Pure GRPO and Fixed
JSD, supporting adaptive JSD weighting based on teacher quality and
all-zero groups.

\begin{table*}[tbp]
\centering
\caption{Effect of the adaptive distillation controller.}
\label{tab:controller_ablation}

\scriptsize
\setlength{\tabcolsep}{2.4pt}
\renewcommand{\arraystretch}{1.0}

\begin{tabular*}{\textwidth}{
@{\extracolsep{\fill}}lcccccccccccc@{}
}
\toprule
& \multicolumn{4}{c}{\textbf{Charades-TimeLens}}
& \multicolumn{4}{c}{\textbf{ActivityNet-TimeLens}}
& \multicolumn{4}{c}{\textbf{QVHighlights-TimeLens}} \\
\cmidrule(lr){2-5}
\cmidrule(lr){6-9}
\cmidrule(lr){10-13}

\textbf{Variant}
& R1@0.3 & R1@0.5 & R1@0.7 & mIoU
& R1@0.3 & R1@0.5 & R1@0.7 & mIoU
& R1@0.3 & R1@0.5 & R1@0.7 & mIoU \\
\midrule

Fixed JSD
& 77.18 & 63.98 & 35.61 & 55.48
& 68.51 & 59.91 & 42.33 & 54.02
& 79.59 & 70.12 & 54.99 & 65.78 \\

Rescue only
& 77.61 & 64.45 & 36.11 & 55.77
& 69.25 & 60.45 & 43.01 & 54.50
& 80.14 & 70.99 & 56.47 & 66.33 \\

Suppression only
& 77.45 & 64.31 & 35.95 & 55.62
& 69.10 & 60.33 & 42.82 & 54.37
& 80.20 & 70.80 & 55.88 & 66.24 \\

Full controller
& \textbf{77.94}
& \textbf{64.67}
& \textbf{36.43}
& \textbf{55.93}
& \textbf{69.67}
& \textbf{60.71}
& \textbf{43.47}
& \textbf{54.77}
& \textbf{80.53}
& \textbf{71.45}
& \textbf{57.39}
& \textbf{66.71} \\

\bottomrule
\end{tabular*}
\end{table*}

\paragraph{Effect of the privileged video design.}
Table~\ref{tab:privilege_ablation} shows that the gains depend on
correct temporal cues rather than colored borders alone. Moving the
red border to an incorrect interval gives 64.44 mIoU, whereas marking
the true target reaches 66.12. Adding random distractor borders
slightly reduces mIoU to 65.94. Replacing them with verified same-video
distractors improves R1@0.7 from 55.68 to 57.39 and mIoU from 65.94
to 66.71. Thus, the improvement depends on both an aligned target cue
and meaningful distractor cues.

\paragraph{Effect of training data curation.}
Table~\ref{tab:data_ablation} compares three 300-example subsets from
the same TimeLens-100K training pool. In the random subset, the
original target is marked in red, while unverified intervals associated
with other queries from the same video serve as candidate blue cues.
Human screening and verification increase mIoU from 65.53 to 66.05,
and teachability filtering further increases it to 66.71. These results
show the cumulative benefits of example selection, distractor
verification, and teacher--student teachability filtering.

\begin{table*}[tbp]
\centering
\begin{minipage}[t]{0.49\textwidth}
\vspace{0pt}%
\centering
\captionof{table}{Privileged-view ablation on QVHighlights-TimeLens.}
\label{tab:privilege_ablation}

\scriptsize
\setlength{\tabcolsep}{2.2pt}
\renewcommand{\arraystretch}{1.0}

\begin{tabularx}{\linewidth}{@{}Xcccc@{}}
\toprule
\textbf{Privileged view}
& R1@0.3 & R1@0.5 & R1@0.7 & mIoU \\
\midrule

Red-shuffled target
& 77.78 & 68.23 & 51.97 & 64.44 \\

Target only
& 80.18 & 71.18 & 56.10 & 66.12 \\

Target + random distractors
& 80.02 & 70.96 & 55.68 & 65.94 \\

Target + verified distractors
& \textbf{80.53}
& \textbf{71.45}
& \textbf{57.39}
& \textbf{66.71} \\

\bottomrule
\end{tabularx}
\end{minipage}
\hfill
\begin{minipage}[t]{0.49\textwidth}
\vspace{0pt}%
\centering
\captionof{table}{Data-curation ablation on QVHighlights-TimeLens.}
\label{tab:data_ablation}

\scriptsize
\setlength{\tabcolsep}{2.2pt}
\renewcommand{\arraystretch}{1.0}

\begin{tabularx}{\linewidth}{@{}Xcccc@{}}
\toprule
\textbf{Training subset}
& R1@0.3 & R1@0.5 & R1@0.7 & mIoU \\
\midrule

Random subset
& 79.75 & 70.62 & 54.90 & 65.53 \\

Human-verified subset
& 80.14 & 71.26 & 56.02 & 66.05 \\

\textsc{HI-TVG-300}
& \textbf{80.53}
& \textbf{71.45}
& \textbf{57.39}
& \textbf{66.71} \\

\bottomrule
\end{tabularx}
\end{minipage}
\end{table*}

\FloatBarrier

\section{Conclusion}

We study same-video interference in TVG, where a model can recognize
the queried event yet select the wrong occurrence. Under strong
interference, all sampled intervals may miss the target, producing an
all-zero tIoU group with no group-relative GRPO signal. DTP-OPSD
addresses this failure by combining GRPO with distractor-aware
temporal-privilege on-policy self-distillation. A shared-weight teacher
uses training-only target and distractor cues to provide token-level
JSD supervision on student-generated prefixes, while an adaptive
controller suppresses supervision when the teacher does not outperform
the student and strengthens it when the teacher performs better but all
student rollout tIoUs are zero. Using only 300 curated examples derived
from TimeLens-100K, DTP-OPSD achieves the highest mIoU among the
compared open-source 8B models on all three TimeLens benchmarks.
Compared with Pure GRPO trained on the same data, it improves all 12
reported metrics, with R1@0.7 gains of 1.28, 1.62, and 1.78 points.
These gains across datasets and overlap thresholds show that
training-only temporal privilege can complement relative-reward
optimization without additional inputs or parameters during inference.

\newpage

\section*{AI Use Statement}
This section is \textbf{required} and does not count toward the page
limit.
In this work, we used generative AI tools for initial candidate screening
during dataset construction and for assistance with manuscript editing,
formatting, and presentation. AI-generated screening outputs were treated
only as candidates and were subsequently verified by human annotators. The
authors reviewed all AI-assisted work, verified the experimental results and
claims, and take full responsibility for the final content of this paper.

\section*{Ethics Statement}
This study uses publicly available research datasets and model
checkpoints; no private or newly collected personal data are involved,
and no identity-related annotations are introduced. We follow the
corresponding licenses and usage guidelines. Although this work poses
no apparent ethical risks, video models may inherit privacy,
demographic, and contextual biases from their source data. Their
outputs may also be misapplied in unforeseen contexts. We therefore
advocate responsible use and recommend additional privacy, fairness,
and safety reviews before deployment in surveillance or other
high-impact applications.

\section*{Reproducibility Statement}
We are committed to ensuring the reproducibility of our work. The
model initialization, training data, evaluation benchmarks, sampling
settings, and hyperparameters are documented in
Section~\ref{sec:experimental_setup}. The method formulation,
training algorithm, prompts, reward parser, temporal-privilege
rendering procedure, and dataset statistics are provided in the main
paper and supplementary material. GPT-5.6 is used only for initial
candidate screening, while the resulting annotations are verified by
human annotators. We plan to release the verified annotations,
training configuration, and evaluation code to support reproduction
of our experiments.

\bibliography{dtp_opsd_references_2027}
\bibliographystyle{iclr2027_conference}

\newpage
\appendix

\setcounter{topnumber}{3}
\setcounter{bottomnumber}{2}
\setcounter{totalnumber}{5}
\renewcommand{\topfraction}{0.92}
\renewcommand{\bottomfraction}{0.85}
\renewcommand{\textfraction}{0.06}
\renewcommand{\floatpagefraction}{0.70}
\setlength{\textfloatsep}{10pt plus 2pt minus 2pt}
\setlength{\floatsep}{8pt plus 2pt minus 2pt}
\setlength{\intextsep}{10pt plus 2pt minus 2pt}

\section{Implementation and Reproducibility Details}
\label{app:implementation_details}

\subsection{Training Objective and Update Procedure}
\label{app:training_algorithm}

For completion token $y_{g,i,t}$, the importance ratio between the
current policy and the rollout snapshot is

\begin{equation}
\rho_{g,i,t}(\theta)
=
\frac{
\pi_{\theta}(y_{g,i,t}\mid V_g,q_g,\hat y^{S}_{g,i,<t})
}{
\pi_{\theta_{\mathrm{old}}}(y_{g,i,t}\mid V_g,q_g,\hat y^{S}_{g,i,<t})
}.
\label{eq:app_importance_ratio}
\end{equation}

Using the group-relative advantage $A_{g,i}$ from
Equation~\ref{eq:group_advantage}, define

\begin{equation}
\widetilde{\rho}_{g,i,t}(\theta)
=
\operatorname{clip}\bigl(
\rho_{g,i,t}(\theta),1-\epsilon,1+\epsilon
\bigr).
\label{eq:app_clipped_ratio}
\end{equation}

The GRPO loss is

\begin{equation}
\begin{aligned}
\mathcal{L}_{\mathrm{GRPO}}
&=
-\frac{1}{|\mathcal{B}|}
\sum_{g\in\mathcal{B}}
\frac{1}{G}
\sum_{i=1}^{G}
\frac{1}{T_{g,i}}
\sum_{t=1}^{T_{g,i}}\\
&\quad
\min\bigl(
\rho_{g,i,t}(\theta)A_{g,i},
\widetilde{\rho}_{g,i,t}(\theta)A_{g,i}
\bigr).
\end{aligned}
\label{eq:app_grpo_loss}
\end{equation}

Here, $\mathcal{B}$ is a batch of prompt-level rollout groups, $G=8$ is
the number of student completions per prompt, and $T_{g,i}$ is the number
of valid non-padding completion tokens. Prompt tokens are excluded, while
an emitted end-of-sequence token is included. Each rollout batch is used
for a single optimizer update. Therefore,
$\theta=\theta_{\mathrm{old}}$ when the gradient is evaluated and the
importance ratio equals one at the update point, making clipping inactive
in our setting. Equation~\ref{eq:app_grpo_loss} consequently produces the
same update direction as the corresponding group-relative policy-gradient
objective. We retain the standard GRPO ratio form to make its relation to
the conventional implementation explicit. The reference-policy KL
coefficient is $\beta=0$, so no reference-KL term is added. The GRPO and
JSD losses use the same completion-level and token-level normalization
before they are combined in Equation~\ref{eq:full_objective}. Algorithm~\ref{alg:training} summarizes the complete training
procedure.

\begin{algorithm}[H]
\small
\caption{\method{} Training with Adaptive Temporal-Privilege
Distillation}
\label{alg:training}
\begin{algorithmic}[1]
\STATE \textbf{Input:} Training set $\mathcal{D}$, shared policy
$\pi_\theta$, rollout size $G$, clipping width $\epsilon$, and global
coefficient $\lambda_{\mathrm{JSD}}$.
\STATE \textbf{Output:} Optimized clean-video policy $\pi_\theta$.
\STATE Pre-render each privileged view $V_g^{\mathrm{priv}}$ by marking
the expanded target interval in red and verified distractors in blue.
\FOR{each prompt batch $\mathcal{B}$ in one training epoch}
    \STATE Set the rollout snapshot
    $\pi_{\theta_{\mathrm{old}}}\leftarrow\pi_\theta$.
    \FOR{each rollout group $g\in\mathcal{B}$}
        \STATE Sample $G$ student responses from
        $\pi_{\theta_{\mathrm{old}}}(\cdot\mid V_g,q_g)$.
        \STATE Parse their intervals, compute student tIoU rewards, and
        obtain the group-relative advantages
        $\{A_{g,i}\}_{i=1}^{G}$.
        \STATE Independently sample $G$ teacher responses from
        $\pi_{\theta_{\mathrm{old}}}
        (\cdot\mid V_g^{\mathrm{priv}},q_g)$ and compute
        $\bar r_g^T$ for the controller.
        \FOR{each student response $\hat y_{g,i}^{S}$ and valid token $t$}
            \STATE Evaluate the student distribution
            $p_{S,g,i}^{t}$ on
            $(V_g,q_g,\hat y_{g,i,<t}^{S})$ with gradients.
            \STATE Evaluate the teacher distribution
            $p_{T,g,i}^{t}$ on
            $(V_g^{\mathrm{priv}},q_g,\hat y_{g,i,<t}^{S})$
            with stop-gradient.
        \ENDFOR
        \STATE Compute the group-level JSD
        $\ell_g^{\mathrm{JSD}}$ and select
        $\lambda_g^{\mathrm{adapt}}\in\{0,1,2\}$ using
        Equation~\ref{eq:quality_controller}.
    \ENDFOR
    \STATE Compute $\mathcal{L}_{\mathrm{GRPO}}$ using
    Equation~\ref{eq:app_grpo_loss}.
    \STATE Form $\mathcal{L}_{\mathrm{total}}$ using
    Equation~\ref{eq:full_objective} and update $\theta$ once.
\ENDFOR
\end{algorithmic}
\end{algorithm}

For distillation, the teacher does not provide a separate target
response. Its independently sampled responses are used only to estimate
the teacher reward for the adaptive controller. The teacher distribution
used by JSD is instead evaluated on each student-generated prefix.

Each rollout batch is used for one optimizer update and is not reused for
inner optimization. Pure GRPO and \method{} follow the same single-update
convention. Teacher distributions are detached, recomputed after every
update, and never cached across updates.

\subsection{Training Protocol}
\label{app:checkpoint_selection}

A global batch contains eight prompt groups, each with eight student
completions, giving 64 student completions per update. \method{} also
samples eight teacher completions per prompt for controller estimation.
Teacher evaluation on student prefixes is a separate forward pass rather
than an additional rollout.
The dataloader drops the final incomplete batch. One epoch over the
300-example set therefore contains 37 optimizer updates, processes 296
examples, and omits four examples from the shuffled order. Within every
paired run, Pure GRPO and \method{} use the same order and omit the same
examples.
Except for the compute-matched control in
Section~\ref{app:compute_comparison}, all methods use one epoch and
report the final checkpoint after 37 updates. We do not select an
intermediate checkpoint based on benchmark performance. Following the
experimental practice of TimeLens~\citep{Zhang2026TimeLens}, we use one
shared configuration across the TimeLens benchmarks. For compactness,
detailed sensitivity results are shown on QVHighlights-TimeLens. The
resulting configuration is applied unchanged to OMTG-Bench and
QV-M$^{2}$.
Unless otherwise specified, results from our post-training experiments
and training ablations are reported as the mean over three independent
training runs. The same three seeds are used for paired method
comparisons. Corresponding standard deviations of the primary mIoU
results are reported in Section~\ref{app:seed_stability}.

\subsection{Memory-Efficient JSD}
\label{app:jsd_implementation}

At a student-prefix position, let $p_S$ and $p_T$ be the clean-video
student and privileged-view teacher distributions. We use symmetric JSD
with equal mixture weights and temperature $1$:

\begin{equation}
q=\operatorname{sg}[p_T],
\qquad
m=\frac{1}{2}(p_S+q),
\label{eq:app_jsd_mixture}
\end{equation}

\begin{equation}
\operatorname{JSD}(p_S,q)
=
\frac{1}{2}\operatorname{KL}(p_S\Vert m)
+
\frac{1}{2}\operatorname{KL}(q\Vert m).
\label{eq:app_jsd_operational}
\end{equation}

The teacher distribution is detached, while the mixture retains its
dependence on $p_S$. Thus, both terms contribute gradients to the student,
but no gradient flows through the teacher branch.

To reduce memory, let $\mathcal{K}_t$ contain the 100 tokens with the
highest teacher probability at position $t$. For either distribution $p$,
we use the deterministic mapping

\begin{equation}
\Phi_t(p)
=
\left(
\{p(v)\}_{v\in\mathcal{K}_t},
\sum_{v\notin\mathcal{K}_t}p(v)
\right).
\label{eq:app_topk_mapping}
\end{equation}

We compute
$\operatorname{JSD}(\Phi_t(p_S),\operatorname{sg}[\Phi_t(p_T)])$ over the
resulting normalized 101-category distributions. The final category keeps
all probability mass outside the teacher top 100, so the retained tokens
are not renormalized in isolation. By the data-processing inequality, this
coarse-grained JSD is a lower bound on the full-vocabulary JSD and provides
a memory-efficient approximation that preserves both the high-probability
tokens and the remaining tail mass. Tokens merged into the tail category
are constrained through their total probability mass rather than their
individual probabilities.

Teacher forward passes use stop-gradient and BF16. Teacher rollout
generation, prefix evaluation, and student optimization are performed
sequentially, and compressed teacher logits are released immediately.
Training uses gradient checkpointing and DeepSpeed
ZeRO-2~\citep{Rajbhandari2020ZeRO} on eight NVIDIA H20 GPUs.

\subsection{Prompt, Reward Parsing, and Video Rendering}
\label{app:reward_prompt}
\label{app:rendering_details}

The parser extracts the predicted start and end times and computes tIoU
with the ground-truth interval. An unparseable interval, a negative
timestamp, or an end time earlier than the start time receives zero
reward. We use no separate format or temporal-distance reward.

The student and teacher use the same prompt:

\begin{quote}
\small
\texttt{Please find the visual event described by the sentence}\\
\texttt{'\{query\}', determining its starting and ending times.}\\
\texttt{The format should be: 'The event happens in}\\
\texttt{<start time> - <end time> seconds'.}
\end{quote}

An expected response is

\begin{quote}
\small
\texttt{The event happens in 12.0 - 18.5 seconds.}
\end{quote}

The prompt does not name the border colors and requires neither a
\texttt{<think>} section nor an \texttt{<answer>} tag.

Privileged videos are pre-rendered and cached. Each target interval is
expanded by $1.5\times$ about its center and clipped to the valid video
range, so that the displayed cue does not directly coincide with the
annotated boundaries. Frames in this region receive red full-frame
borders, while frames in verified distractor intervals receive blue
borders; red takes priority in overlap regions. Distractor intervals
are not expanded, and the student receives the unchanged video.

\subsection{Training Set and Evaluation Protocol}
\label{sec:data_stats}
\label{app:evaluation_protocol}

\textsc{HI-TVG-300} contains 300 query--video pairs from the
TimeLens-100K training split, without external videos or queries. Each
example has at least two human-verified distractor occurrences from the
same video. Candidates are sampled across source datasets and duration
ranges. A VLM proposes repeated, similar, related, or opposite
occurrences, after which human annotators verify the video, query, target
boundaries, and all distractor intervals.

Using the same Qwen3-VL-8B checkpoint and sampling settings, the clean
student and privileged teacher each generate eight independent
completions during offline filtering. We retain an example when the
teacher mean tIoU is above $0.7$ and the student mean tIoU is below $0.3$.
Released annotations include the refined target and all verified
distractor intervals.

The human-verified control in Table~\ref{tab:data_ablation} is sampled
after VLM screening and manual verification but before the
teacher--student teachability filter. It therefore uses refined target
boundaries and verified distractor intervals without filtering examples
according to the privileged-teacher advantage.

All methods in the main comparison are evaluated using the corrected
TimeLens annotations and the same TimeLens evaluation protocol. Results
from our implementation use the same prompt, parser, and tIoU code. Videos
are sampled at 2 FPS with at most 448 frames and a visual-token budget of
14,336. For QV-M$^{2}$, we evaluate all 323 queries in the official test
split and apply no availability-based query filtering.

For the distractor-count analysis in Figure~\ref{fig:motivation}, target
windows come from the corrected evaluation annotations. Candidate
same-video distractors are screened and manually verified before examples
are grouped by distractor count. These evaluation-side annotations are
used only for analysis and do not affect training or model selection.

The random control is a fixed 300-example draw from the same
TimeLens-100K pool. For each query, its original TimeLens target interval
is expanded by $1.5\times$ and marked in red. Intervals annotated for
other queries associated with the same video are used as candidate blue
cues when available. This row uses neither manual target-boundary
refinement nor human verification of whether the blue intervals are valid
distractors for the current query. It obtains $79.75/70.62/54.90$ for
R1@0.3/0.5/0.7 and $65.53$ mIoU on QVHighlights-TimeLens. All rows use
\method{}, which isolates the effect of successive stages in the
data-curation pipeline under a fixed training method.

\section{Additional Ablations and Sensitivity}
\label{app:additional_ablations}

\subsection{Contribution of the Training Objectives}
\label{app:objective_components}

All trained variants in Tables~\ref{tab:objective_components} and
\ref{tab:gt_sft} use the same initialization, training data, one-epoch
budget, and evaluation protocol. Rows containing privileged JSD use the
adaptive controller in Equation~\ref{eq:quality_controller}; the No/No
row is the TimeLens-SFT initialization without additional training.

\begin{table}[htbp]
\centering
\begin{minipage}[t]{0.49\textwidth}
\vspace{0pt}
\centering
\captionof{table}{Objective components on QVHighlights-TimeLens.}
\label{tab:objective_components}
\small
\begin{tabular}{llc}
\toprule
GRPO & Privileged JSD & mIoU \\
\midrule
No  & No  & 57.34 \\
Yes & No  & 65.95 \\
No  & Yes & 15.04 \\
Yes & Yes & \textbf{66.71} \\
\bottomrule
\end{tabular}
\end{minipage}
\hfill
\begin{minipage}[t]{0.49\textwidth}
\vspace{0pt}
\centering
\captionof{table}{Comparison with ground-truth SFT on
QVHighlights-TimeLens.}
\label{tab:gt_sft}
\small
\begin{tabular}{lc}
\toprule
Method & mIoU \\
\midrule
Pure GRPO & 65.95 \\
GRPO + GT-SFT & 66.23 \\
\method{} & \textbf{66.71} \\
\bottomrule
\end{tabular}
\end{minipage}
\end{table}

For rows containing privileged JSD, student and teacher rollout rewards
are retained for controller estimation even when the GRPO loss is
disabled. Privileged JSD is a consistency objective rather than a
sequence-level correctness objective. GRPO anchors optimization to the
task-level tIoU reward, while JSD provides token-level guidance on
student-generated prefixes. The JSD-only result shows that privileged JSD
does not provide a sufficient standalone training objective. Combining it
with GRPO increases mIoU from 65.95 to 66.71, suggesting that the two
objectives are complementary in this setting.

\subsection{Comparison with Ground-Truth SFT}
\label{app:gt_sft_comparison}

We compare \method{} with a standard direct-supervision alternative.
GRPO + GT-SFT jointly optimizes the GRPO objective and a token-level
cross-entropy loss on the ground-truth timestamp response for every
training example. The SFT coefficient is set to $1$. All methods use the
same initialization, training data, one-epoch budget, and evaluation
protocol. GRPO + GT-SFT reaches 66.23 mIoU, compared with 65.95 for
Pure GRPO. Under the same setting, \method{} reaches 66.71 mIoU,
compared with 66.23 for GRPO + GT-SFT.

\subsection{Global JSD Weight}
\label{app:jsd_weight_sensitivity}
\begin{figure}[htbp]
\centering
\includegraphics[
    width=\linewidth,
    keepaspectratio
]{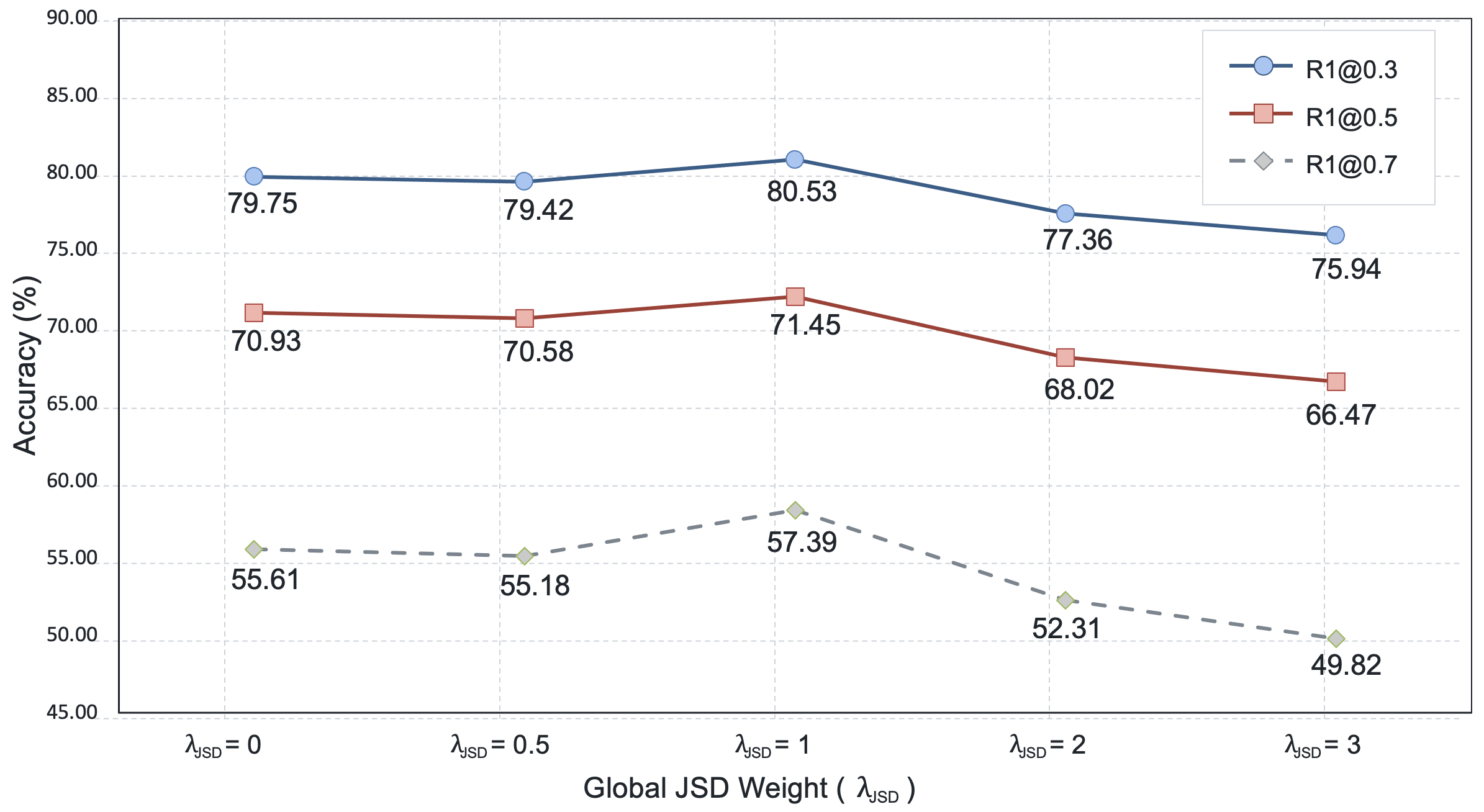}
\caption{R1 on QVHighlights-TimeLens under different global JSD
coefficients. The adaptive controller is enabled for every nonzero
coefficient, and the final configuration uses
$\lambda_{\mathrm{JSD}}=1$. Values are means over three training runs.}
\label{fig:jsd_weight_sensitivity}
\end{figure}

We use $\lambda_{\mathrm{JSD}}=1$ as the shared value across all
three TimeLens benchmarks. This choice is not based solely on
QVHighlights-TimeLens. For compactness,
Figure~\ref{fig:jsd_weight_sensitivity} reports only
QVHighlights-TimeLens, where all three R1 metrics peak at $1$ and
decrease at larger weights.

\subsection{Target Expansion and Dead-Group Recovery}
\label{app:target_expansion}
\label{app:dead_group_diagnostics}

\paragraph{Target expansion.}
We compare three variants that differ only in the duration of the
displayed target region. The $1.0\times$ setting displays the annotated
boundaries, whereas larger ratios provide progressively coarser temporal
cues. Table~\ref{tab:target_expansion} reports the mean results over the
same three training runs. Performance varies by at most 0.35 mIoU across
the tested ratios, and we use $1.5\times$ throughout the main experiments.

\paragraph{Dead-group recovery.}
Using a fixed diagnostic rollout configuration, 142 of the 300 training
examples form all-zero groups at initialization: all eight student
completions have zero tIoU. We evaluate this fixed subset at the three
final checkpoints using the same diagnostic rollout configuration and
mark an example as recovered when at least one completion has positive
tIoU. The reported results are averaged over the three checkpoints, and
all percentages use the same 142 examples as the denominator. The recovery results are reported in
Table~\ref{tab:dead_group_recovery}.

\paragraph{Destination of initial dead-group predictions.}
We further examine all $142\times8=1{,}136$ student rollouts from the
142 initial dead groups. For each prediction, we compute its maximum
tIoU with all human-verified distractor intervals in the same video.
A prediction is counted as distractor-overlapping when this maximum
tIoU is greater than zero. All 1,136 predictions have zero tIoU with
the target interval by construction.
Table~\ref{tab:dead_group_destination} summarizes the resulting
overlap outcomes.

\begin{table}[htbp]
\centering
\small
\caption{
Destination of predictions from the 142 initial dead groups.
Percentages are computed over all 1,136 student rollouts.
}
\label{tab:dead_group_destination}
\begin{tabular}{lc}
\toprule
Prediction outcome & Percentage \\
\midrule
Overlaps at least one verified distractor & \textbf{79.8\%} \\
No verified distractor overlap & 20.2\% \\
\bottomrule
\end{tabular}
\end{table}

The high distractor-overlap rate indicates that many all-zero rollouts
intersect verified competing occurrences rather than the annotated
target. This pattern suggests that incorrect occurrence selection is a
common source of all-zero groups in this diagnostic subset.

\begin{table}[htbp]
\centering
\begin{minipage}[t]{0.34\textwidth}
\vspace{0pt}
\centering
\captionof{table}{Target-expansion sensitivity on
QVHighlights-TimeLens.}
\label{tab:target_expansion}
\small
\begin{tabular}{@{}lc@{}}
\toprule
Expansion ratio & mIoU \\
\midrule
$1.0\times$ & 66.51 \\
$1.5\times$ & \textbf{66.71} \\
$2.0\times$ & 66.36 \\
\bottomrule
\end{tabular}
\end{minipage}
\hfill
\begin{minipage}[t]{0.62\textwidth}
\vspace{0pt}
\centering
\captionof{table}{Recovery of the fixed initial all-zero subset at the
final checkpoints.}
\label{tab:dead_group_recovery}

\scriptsize
\renewcommand{\arraystretch}{1.45}
\begin{tabular}{@{}lrrr@{}}
\toprule
Method & Initial & Recovered & At least one tIoU $\geq 0.3$ \\
\midrule
Pure GRPO & 142 & 55 (38.7\%) & 27.5\% \\
\method{} & 142 & \textbf{91 (64.1\%)} & \textbf{48.6\%} \\
\bottomrule
\end{tabular}
\end{minipage}
\end{table}

\method{} recovers more examples that initially provide no
group-relative policy gradient, providing a targeted diagnostic of the
failure mode addressed by the method.

\FloatBarrier
\section{Additional Qualitative Example}
\label{sec:qualitative_examples}

\begin{figure}[H]
\centering
\includegraphics[width=\textwidth]{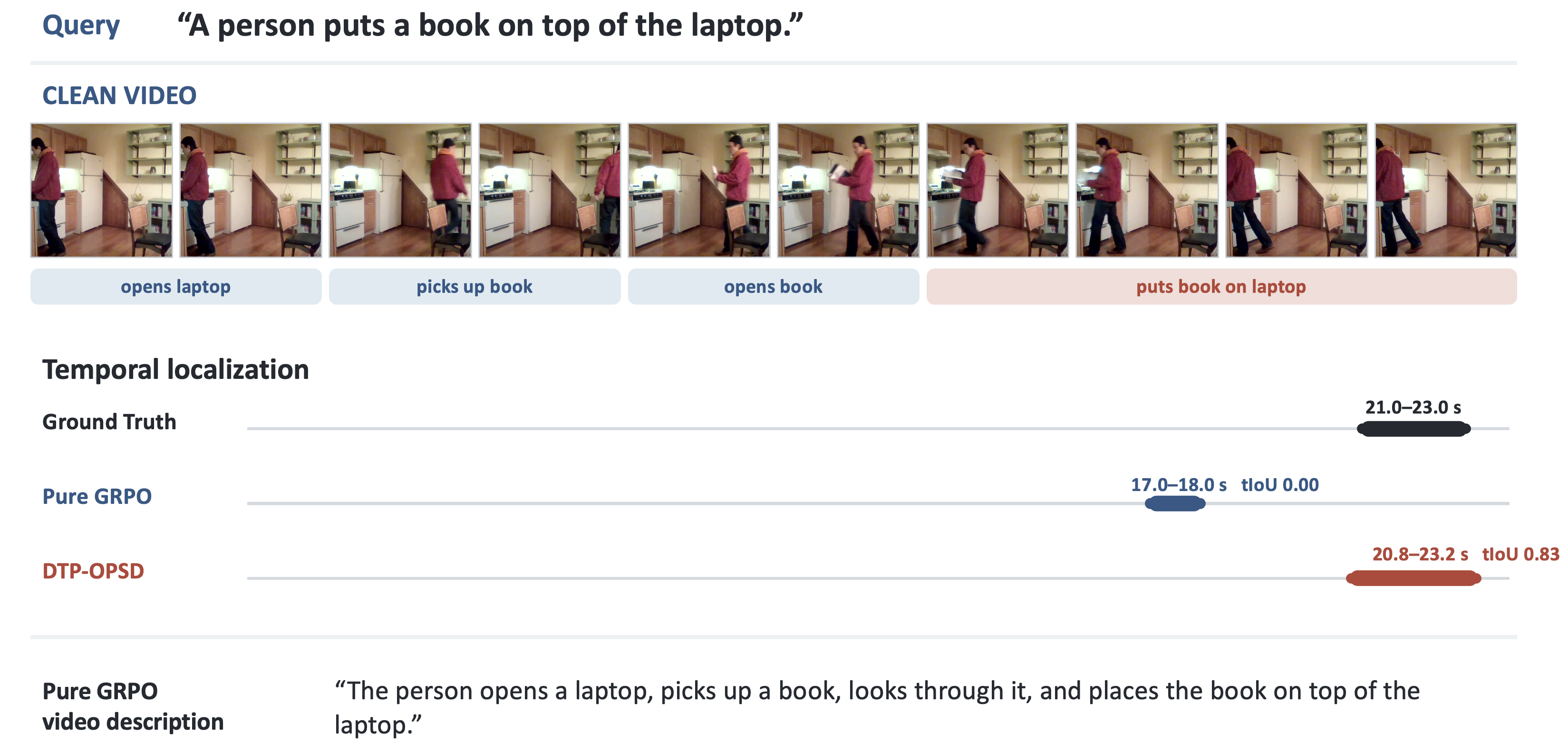}
\caption{Qualitative comparison on a video with related actions involving
a laptop and a book. Pure GRPO selects an earlier occurrence, while
\method{} recovers the queried occurrence and predicts close boundaries.}
\label{fig:qualitative_case_sp1}
\end{figure}

Figure~\ref{fig:qualitative_case_sp1} shows a same-video interference
case. Pure GRPO recognizes the relevant objects and action but predicts
$17.0$--$18.0$ seconds, which has no overlap with the ground-truth
interval of $21.0$--$23.0$ seconds. \method{} predicts
$20.8$--$23.2$ seconds and reaches a tIoU of $0.83$. The improvement
comes from selecting the correct occurrence rather than adjusting the
boundaries around the same incorrect occurrence.

\FloatBarrier

\section{Stability and Efficiency}
\label{app:training_efficiency}

\subsection{Stability Across Seeds}
\label{app:seed_stability}

Table~\ref{tab:seed_stability} reports the mean and sample standard
deviation of mIoU over the same three paired training runs. Within each
seed pair, Pure GRPO and \method{} use the same initialization, training
data order, and student-rollout seed. Both methods have standard
deviations below 0.12 mIoU on all three benchmarks, while \method{}
achieves the higher mean in each case.

\begin{table}[htbp]
\centering
\small
\caption{Mean mIoU and sample standard deviation over three paired
training runs.}
\label{tab:seed_stability}
\begin{tabular}{lccc}
\toprule
Method & Charades-TimeLens & ActivityNet-TimeLens &
QVHighlights-TimeLens \\
\midrule
Pure GRPO
& $55.43\pm0.08$
& $53.62\pm0.10$
& $65.95\pm0.11$ \\
\method{}
& $\mathbf{55.93\pm0.06}$
& $\mathbf{54.77\pm0.07}$
& $\mathbf{66.71\pm0.09}$ \\
\bottomrule
\end{tabular}
\end{table}

\subsection{Compute-Matched Comparison}
\label{app:compute_comparison}

We measure wall time from the first rollout to the final update. Offline
privileged-video rendering takes 0.6 CPU-hours and is excluded from
training GPU-hours. GPU-hours equal wall time multiplied by eight GPUs.
The compute-matched Pure GRPO run is trained continuously to approximately
the same GPU-hour budget as \method{}, and its final checkpoint is reported
without intermediate selection.

As shown in Table~\ref{tab:compute_comparison}, additional GRPO
computation improves the one-epoch baseline but does not match
\method{}. Under a comparable GPU-hour budget, compute-matched Pure
GRPO reaches 66.09 mIoU, compared with 66.71 for \method{}. Under this
matched budget, longer GRPO training narrows but does not close the
observed performance gap.

\begin{table}[htbp]
\centering
\small
\caption{Training cost on eight NVIDIA H20 GPUs and accuracy on
QVHighlights-TimeLens. Peak memory is reported per GPU. Values are
averaged over the same three runs.}
\label{tab:compute_comparison}
\begin{tabular}{lrrrr}
\toprule
Method & Wall time & GPU-hours & Peak memory & mIoU \\
\midrule
Pure GRPO, one epoch
& 7.3 h & 58.4 & 54.6 GB & 65.95 \\
Pure GRPO, compute matched
& 11.7 h & 93.6 & 54.6 GB & 66.09 \\
\method{}, one epoch
& 11.7 h & 93.6 & 59.1 GB & \textbf{66.71} \\
\bottomrule
\end{tabular}
\end{table}

\section{Results Across Model Scales}
\label{app:model_scaling}

We evaluate \method{} on the 2B, 4B, and 8B Qwen3-VL backbones. All scales
use the same \textsc{HI-TVG-300} subset selected by the 8B model, without
model-specific filtering or example replacement. The teachability
thresholds, JSD coefficient, rescue multiplier, and other method-specific
settings are unchanged. Before training, 61.7\% and 72.3\% of the
8B-selected examples still meet the original teachability rule under the
2B and 4B models, respectively. Table~\ref{tab:model_scaling} reports the corresponding results.

\begin{table}[htbp]
\centering
\small
\caption{Results across model scales using the same 8B-filtered
\textsc{HI-TVG-300} set. Each fixed SFT initialization is evaluated
once, while post-training results are averaged over three runs.
Comparisons are made within each scale.}
\label{tab:model_scaling}
\begin{tabular}{llrrr}
\toprule
Backbone & Method & Charades mIoU & ActivityNet mIoU &
QVHighlights mIoU \\
\midrule
Qwen3-VL-2B & SFT initialization & 42.18 & 41.37 & 54.62 \\
Qwen3-VL-2B & Pure GRPO & 43.36 & 42.11 & 55.80 \\
Qwen3-VL-2B & \method{} & \textbf{43.91} & \textbf{42.68} &
\textbf{56.51} \\
\midrule
Qwen3-VL-4B & SFT initialization & 47.02 & 46.11 & 56.08 \\
Qwen3-VL-4B & Pure GRPO & 48.21 & 47.03 & 60.36 \\
Qwen3-VL-4B & \method{} & \textbf{48.79} & \textbf{47.66} &
\textbf{61.14} \\
\midrule
Qwen3-VL-8B & SFT initialization & 51.85 & 49.26 & 57.34 \\
Qwen3-VL-8B & Pure GRPO & 55.43 & 53.62 & 65.95 \\
Qwen3-VL-8B & \method{} & \textbf{55.93} & \textbf{54.77} &
\textbf{66.71} \\
\bottomrule
\end{tabular}
\end{table}

\method{} improves over the corresponding SFT initialization and Pure GRPO
across the tested model scales. Because the SFT initializations are
trained separately, we focus on within-scale comparisons rather than
monotonic trends across backbone sizes.

\section{Limitations}

\method{} focuses on same-video interference. \textsc{HI-TVG-300}
contains only 300 examples from TimeLens-100K and does not cover every
temporal grounding error, video domain, or annotation style. All tested
backbones also belong to the Qwen3-VL family.
The privileged view requires verified target and distractor intervals,
which adds data-preparation cost. Teachability filtering favors examples
with a clear privileged-teacher advantage and may underrepresent cases in
which both views remain uncertain. Expanding the target cue avoids
directly displaying the annotated boundaries but does not remove all
temporal information leakage.
Teacher rollouts and prefix evaluation increase post-training cost. The
method also remains more expensive to train than Pure GRPO. The controller
relies on Monte Carlo estimates from eight student and eight teacher
completions, so its group-level decisions remain subject to sampling
variance.
The data ablation changes the training subset while keeping \method{}
fixed, so it measures the contribution of the complete curation pipeline
rather than independently isolating every curation stage. The privileged
cues, teacher branch, and adaptive controller are used only during
training. The final model receives clean videos and adds no parameters or
computation at inference.

\end{document}